\documentclass{article}
\usepackage{aaaim}
\usepackage{latexsym}
\usepackage{epsfig}

\begin{document}

\newtheorem{prop}{Proposition}
\newtheorem{defin}{Definition}
\newtheorem{theorem}{Theorem}
\newtheorem{corollary}{Corollary}
\newtheorem{lemma}{Lemma}
\newtheorem{obs}{Observation}

\def\remainder{{\mbox{$\perp$}}}
\def\falsum{{\mbox{{$\scriptstyle\bf\bot$}}}}
\def\kernel{{\mbox{$\;\perp\!\!\!\perp\,$}}}
\def\contract{{\mbox{$\dot{-}$}}}
\def\real{{\mbox{$I\!\!{R}$}}}

\long\def\BC#1\EC{}
\def\EC{}

\title{\bf Local Diagnosis}
\author{Renata Wassermann\\
Institute for Logic, Language and Computation\\
University of Amsterdam\\
email: renata@wins.uva.nl}

\maketitle

\begin{abstract}
In \cite{Local,Frontiers}, we have presented operations of belief change
which only affect the relevant part of a belief base. In this paper, we 
propose the application of the same strategy to the problem of model-based
diangosis. We first isolate the subset of the system description which is
relevant for a given observation and then solve the diagnosis problem for 
this subset. 
\end{abstract}

\section{Introduction}

In \cite{KR00}, we have shown how consistency-based diagnosis relates to
belief revision and how Reiter's algorithm can be used for kernel operations
of belief change. In \cite{Local}, we presented the idea of local change and
characterized operations of belief change that only affect the relevant part
of a belief base. In \cite{Frontiers}, we presented an algorithm for
retrieving the relevant part of a belief base which can be used for
implementing local change. In the present paper, we close the circle by 
showing how local change can be used for focusing the diagnosis process on 
the relevant part of the domain.

We will show how a diagnosis problem can be translated into an operation
of kernel semi-revision. Kernel semi-revision \cite{Hansson_semi}
consists in adding new information to a database and restoring consistency 
if necessary. To restore consistency, the expanded  database is contracted by 
$\falsum$. 


Then we will show how to use information about the structure of the device
being examined in order to obtain more efficient methods of diagnosis.
For this, we will use the  operation of local kernel semi-revision,
presented in \cite{Local}, that considers only the relevant part 
of the database. In \cite{Frontiers}, we have presented a simple
method for extracting the relevant part of a structured database, which 
will be used in this paper.


This paper proceeds as follows: in the next two sections we give a brief
introduction to kernel operations of belief change and consistency-based
diagnosis. Then we show the relation between kernel semi-revision and Reiter
diagnosis. Using this relation, we show how to use information about the 
system to focus on its relevant part during the process of diagnosis.

In the rest of this paper we consider $L$ to be a propositional language
closed under the usual truth-functional connectives and containing a constant
$\falsum$ denoting falsum.

\section{Kernel Semi-Revision}

Hansson introduced a construction for contraction operators, called 
{\it kernel contraction} \cite{Hansson_Kernel}, which is a generalization 
of the operation of safe contraction defined in
\cite{Alchourron_Makinson2}.
The idea behind kernel contraction is that, if we remove from the belief
base $B$ at least one element of each $\alpha$-kernel (minimal subset of
$B$ that implies $\alpha$), then we obtain a belief base that does not
imply $\alpha$ \cite{Hansson_Kernel}. To perform these removals of elements, 
we use an incision function, i.e., a function that selects at least one
sentence from each kernel.

\begin{defin}\cite{Hansson_Kernel}
\label{defkernel}
The {\em\bf kernel} operation $\kernel$ is the operation such that for every
set $B$ of formulas and every formula $\alpha$, $X\in B\kernel\alpha$ if and
only if:
\begin{enumerate}
\item $X\subseteq B$
\item $\alpha\in Cn(X)$
\item for all $Y$, if $Y\subset X$ then $\alpha\not\in Cn(Y)$ 
\end{enumerate}
The elements of $B\kernel\alpha$ are called $\alpha$-kernels.
\end{defin}

\begin{defin} \cite{Hansson_Kernel}
\label{defincision}
An {\em\bf incision function} for $B$ is any function $\sigma$ such that for
any formula $\alpha$:
\begin{enumerate}
\item $\sigma(B\kernel\alpha)\subseteq\bigcup(B\kernel\alpha)$, and
\item If $\emptyset\neq X\in B\kernel\alpha$, then
$X\cap\sigma(B\kernel\alpha)\neq\emptyset$.
\end{enumerate}
\end{defin}

Semi-revision consists in adding new information to a database and restoring
consistency  if necessary. To restore consistency, the expanded  database is
contracted by  $\falsum$. Semi-revision consists of two steps: first
the belief $\alpha$ is added to the base, and then the resulting base is
consolidated, i.e., contracted by $\falsum$. Kernel semi-revision uses kernel
contraction for the second step. 

\begin{defin}
\cite{Hansson_semi}
The  {\em\bf kernel semi-revision} of $B$ based on
an incision function $\sigma$ is the operator $?_\sigma$
such that for all sentences $\alpha$: 

$B?_\sigma \alpha=
(B\cup\{\alpha\})\setminus\sigma((B\cup\{\alpha\})\kernel\falsum)$
\end{defin}

\section{Consistency-Based Diagnosis}

Diagnosis is a very active area within the artificial intelligence
community. The problem of diagnosis consists in, given an observation of an
abnormal behavior, finding the components of the system that may have caused 
the abnormality \cite{Reiter}.


In the area known as model-based diagnosis \cite{Readings_Diagnosis}, 
a model of the device to be diagnosed is given in some 
formal language. In this paper, we will concentrate on model-based diagnosis 
methods that work by trying to restore the consistency of the system
description and the observations.

Although Reiter's 
framework is based on first-order logic, most of the problems studied in 
the literature do not make use of full  first-order logic and can be easily
represented in a propositional language.  For the sake of simplicity, we will
adapt the definitions given in \cite{Reiter} to only mention formulas in  the
propositional language $L$.

\subsection{Basic Definitions}

The systems to be diagnosed will be described by a set of propositional
formulas. For each component $X$ of the system, we use a propositional
variable of the form $okX$ to indicate whether the component is working as
it should. If there is no evidence that the system is not working, we can
assume that variables of the form $okX$ are true.

\begin{defin}
A {\em\bf system} is a pair (SD,ASS), where:
\begin{enumerate}
\item SD, the system description, is a finite set of  formulas of $L$ and
\item ASS, the set of assumables, is a finite set of propositional variables
of the form $okX$.
\end{enumerate} 
\end{defin}

An {\em\bf observation} is a formula of $L$. We will sometimes represent a 
system by (SD,ASS,OBS), where OBS is an observation  for the system (SD,ASS).

The need for a diagnosis arises when an abnormal behavior is observed, i.e.,
when SD$\cup$ASS$\cup$\{OBS\} is inconsistent. A diagnosis is a minimal set of 
assumables that must be negated in order to restore consistency.

\begin{defin}
\label{diag1}
A {\em\bf diagnosis} for (SD,ASS,OBS) is a minimal set $\Delta\subseteq$ASS 
such that:

SD $\cup$ \{OBS\} $\cup$ ASS$\setminus\Delta$ $\cup$ \{$\neg okX |okX
\in\Delta$\} is consistent.
\end{defin}

A diagnosis for a system does not always exist:

\begin{prop}\cite{Reiter}
\label{existdiag}
A diagnosis exists for (SD,ASS,OBS) if and only if SD$\cup$\{OBS\} is
consistent.
\end{prop}

Definition \ref{diag1} can be simplified as follows:

\begin{prop}\cite{Reiter}
\label{proReiter1}
The set $\Delta\subseteq$ASS is a diagnosis for (SD,ASS,OBS) if and only if
$\Delta$ is a minimal set such that
SD $\cup$ \{OBS\} $\cup$ (ASS$\setminus\Delta$) is consistent.
\end{prop}

\subsection{Computing Diagnoses}

In this section we will present Reiter's construction for finding diagnoses.
Reiter's method for computing diagnosis makes use of the concepts of {\it
conflict sets} and {\it hitting sets}. A conflict set is a set of assumables
that cannot be all true given the observation:

\begin{defin}\cite{Reiter}
\label{defconflict}
A {\em\bf conflict set} for (SD,ASS,OBS) is a set
Conf $=\{okX_1,$ $okX_2,$ $...,$ $okX_n\}$ $\subseteq$ ASS such that
SD $\cup$ \{OBS\} $\cup$ Conf is inconsistent. 
\end{defin}

From Proposition \ref{proReiter1} and Definition \ref{defconflict} it follows
that 
$\Delta\subseteq$ASS is a diagnosis for (SD,ASS,OBS) if and only if
$\Delta$ is a minimal set such that ASS$\setminus\Delta$ is not a conflict 
set for (SD,ASS,OBS).

A hitting set for a collection of sets is a set that intersects all sets of
the collection:

\begin{defin}\cite{Reiter}
Let $\cal C$ be a collection of sets. A {\em\bf hitting set} for $\cal C$ is a
set $H\subseteq\bigcup_{S\in\cal C}S$ such that for every $S\in\cal C$, $H\cap
S$ is nonempty.
A hitting set for $\cal C$ is minimal if and only if no proper subset of it is
a hitting set for $\cal C$.
\end{defin}

The following theorem presents a constructive approach for finding diagnoses:

\begin{theorem}\cite{Reiter}
$\Delta\subseteq$ASS is a diagnosis for (SD,ASS,OBS) if and only if
$\Delta$ is a minimal hitting set for the collection of minimal conflict sets
for (SD,ASS,OBS).
\end{theorem}

\begin{figure}[h]
\vspace*{3cm}
\hspace*{2cm}
\begin{picture}(100,100)

 \put(30,30){\mbox{\makebox(100,100)[bl]{\includegraphics{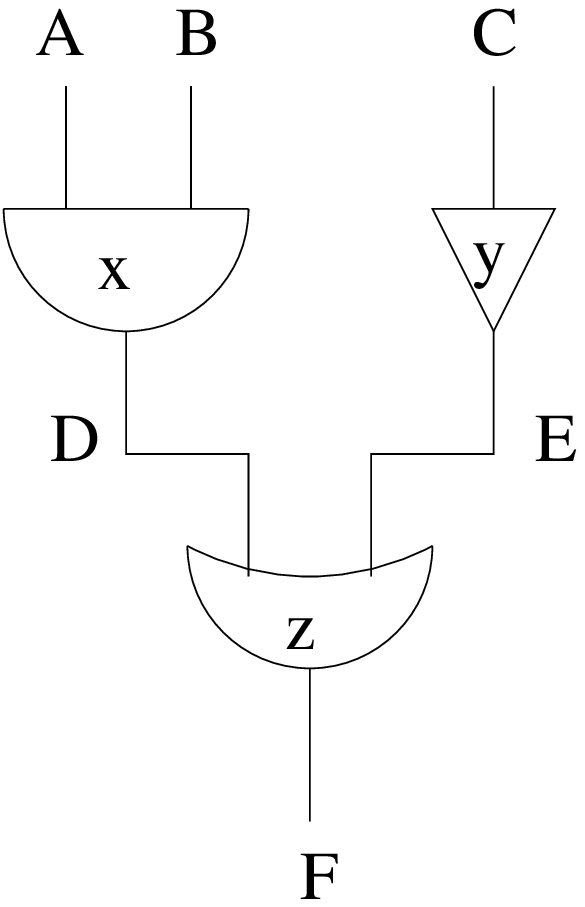}}}}

\end{picture}
\caption{Circuit}
\label{fig6}
\end{figure}

Consider the circuit in Figure \ref{fig6}. The system
description of this circuit is given by (SD,ASS),
where:

\medskip

$\mbox{ASS} = \{okX, okY, okZ\}$

$\begin{array}{ll}\mbox{SD} = &\{(A\wedge B)\wedge okX\rightarrow D,\\
       &\neg(A\wedge B)\wedge okX\rightarrow\neg D,\\
      & C\wedge okY\rightarrow \neg E,\\
      & \neg C\wedge okY\rightarrow E,\\
      & (D\vee E)\wedge okZ\rightarrow F,\\
       &\neg(D\vee E)\wedge okZ\rightarrow \neg F\}\end{array}$

\medskip

Suppose we have OBS=$\neg C\wedge\neg F$. This observation is inconsistent with
$\mbox{SD}\cup\mbox{ASS}$. There is only one minimal conflict set for
(SD,ASS,OBS): \{$okY,okZ$\}.
There are three possible hitting sets: \{$okY$\},\{$okZ$\}, and \{$okY,okZ$\}.
Reiter considers only minimal hitting sets as diagnoses, that is, either
Y or Z is not working well.

\section{Diagnosis via Kernel Semi-Revision}

In \cite{KR00}, we have shown that the standard method for finding
consistency-based diagnosis, due to Reiter \cite{Reiter}, is very similar to
the construction of kernel semi-revision, except for the fact that Reiter 
only considers minimal diagnosis, which correspond to minimal values for 
incision functions. In this section we summarize these results.

Recall that kernel operations are based on two concepts: kernels and incision
functions. The kernels are the minimal subsets of a belief base implying some
sentence, while the incision functions are used to decide which elements of
the kernels should be given up. Let (SD,ASS,OBS) be a system.
The belief base that we are going to semi-revise  corresponds to SD$\cup$ASS
and the input sentence is OBS. The conflict sets are the assumables
in the inconsistent kernels of SD$\cup$ASS$\cup$\{OBS\}. So, if 
$B$=SD$\cup$ASS, the conflict sets are given by \{$X\cap$ASS$|X\in$
($B+$OBS)$\kernel\falsum$\}. Incision functions  correspond loosely to 
hitting sets, the minimal hitting sets being the values of minimal incisions
that return only assumables. Note that there is a difference in the status of
formulas in SD and those in ASS: formulas in ASS represent expectations and
are more easily retracted than those in SD (cf. Definition \ref{modincision}).

We can model the diagnosis problem as a kernel semi-revision by the
observation. Semi-revision can be divided in two steps. First the observation
is added to the system description together with the assumables.  
In case the observation is consistent with the system description together 
with the assumables, no formula has to be given up. Otherwise, we take the 
inconsistent kernels and use an incision function to choose which elements 
of the kernels should be given up.

In the case of diagnosis, we do not wish to give up sentences belonging to the
system description or the observation. We prefer to give up the formulas
of the form $okX$, where $X$ is a component of the system. Moreover, we are 
interested in minimal diagnosis, so the incision should be minimal. For this,
we use a special variant of incision function. We modify Definition
\ref{defincision} so that incisions are minimal and elements of a given set
$A$ are prefered over the others:

\begin{defin}
\label{modincision}
Given a set $A$, an {\em\bf $A$-minimal incision function} is any function
$\sigma_A$ from sets of sets of formulas  into sets of formulas such that for
any set $S$ of sets of formulas:
\begin{enumerate}
\item $\sigma_A(S)\subseteq\bigcup S$, 
\item If $\emptyset\neq X\in S$, then $X\cap\sigma_A(S)\neq\emptyset$, 
\item If  for all  $X\in S$, $X\cap A\neq\emptyset$, then
$\sigma_A(S)\subseteq A$, and
\item $\sigma_A(S)$ is a minimal set satisfying 1,2, and 3.
\end{enumerate}
\end{defin}

If we take $A$ to be the set of assumables, we obtain an incision function
that prefers to select formulas of the form $okX$ over the others.

We can show that for (SD,ASS,OBS), whenever a diagnosis exists, an ASS-minimal
incision function will select only elements of ASS:

\begin{prop}
\label{propmod}
Let (SD,ASS,OBS) be a system with an observation and $\sigma_{ASS}$
an ASS-minimal incision function. If a diagnosis exists, then 
$\sigma_{ASS}((\mbox{SD}\cup\mbox{ASS}\cup\mbox{OBS})\kernel\falsum)\subseteq$ASS. 
\end{prop}

\begin{lemma}
\label{lemmaASS}
The assumables that occur in an inconsistent kernel of the set
SD$ \cup$ASS$ \cup$OBS form a conflict set for (SD,ASS,OBS) and all minimal
conflict sets can be obtained in this way, i.e.:

(i) For every $X\in($SD$ \cup$ASS$ \cup$OBS$)\kernel\falsum$, $X\cap$ASS is a
conflict set, and 

(ii) For every minimal conflict set $Y$, there is some $X\in($SD$ \cup$ASS$
\cup$OBS$)\kernel\falsum$ such that $X\cap$ASS$=Y$.
\end{lemma}

Note that not every inconsistent kernel determines a minimal conflict set,
since for conflict sets only the elements of ASS matter, i.e., there may be
two inconsistent kernels $X_1$ and $X_2$ such that $X_1\cap$ASS is a proper
subset of $X_2\cap$ASS.

Recall that given an incision function
$\sigma$, the semi-revision of a set $B$ by a formula $\alpha$ was given by
$B?_\sigma\alpha=(B+\alpha)\setminus\sigma((B+\alpha)\kernel\falsum)$.
A diagnosis is given by the elements of ASS that are given up in a kernel
semi-revision by the observation.

\begin{prop}
\label{diag2}
Let S=(SD,ASS,OBS) be a system and $\sigma_{ASS}$ an ASS-min\-i\-mal incision
function.

(SD $\cup$ASS)$\setminus$((SD $\cup$ASS)$?_{\sigma_{ASS}}$OBS) =
$\sigma_{ASS}$((SD $\cup$ASS $\cup$OBS)$\kernel\falsum$) is a diagnosis.

\end{prop} 

\section{Using System Structure}

\begin{figure}[h]
\vspace*{6cm}
\hspace*{-1cm}
\begin{picture}(100,100)

 \put(30,30){\mbox{\makebox(100,100)[bl]{\includegraphics{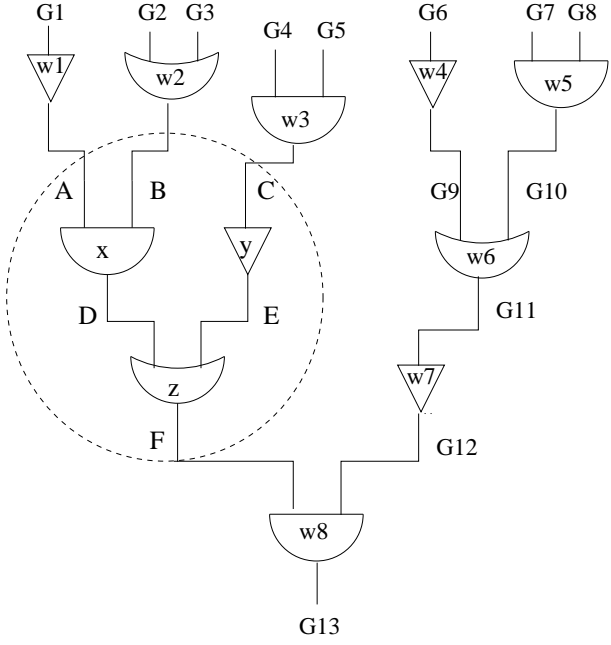}}}}

\end{picture}
\caption{Larger Circuit}
\label{fig7}
\end{figure}

Suppose that instead of the circuit depicted in Figure \ref{fig6}, we have the
circuit in Figure \ref{fig7}. Suppose also that we get the same observation,
i.e., OBS$=\neg C\wedge\neg F$. Intuitively, only a small part of the
circuit (roughly the sub-circuit at figure \ref{fig6}) has to be  considered
in order to arrive to a diagnosis.

In \cite{Local}, we have extended the definition of kernel semi-revision to an
operation that considers only the relevant part of a database, local kernel
semi-revision. In \cite{Frontiers}, we have shown how to use structure present
in a database in order to find compartments and implement local kernel
operations more efficiently. The key idea of the method described is to use a
relation of relatedness between formulas of the belief base. In some
applications, as we will see, such a relation is given with the problem.
In the case of the circuit shown in Figure \ref{fig7}, there
is a very natural dependence relation. The output of each of the components
depends on the input and on whether the component  is working well. 

The only observation we have is $\neg C\wedge\neg F$. Since this observation is
inconsistent with the system description together with the assumption that all
components are working well, there must be some faulty component. Moreover,
the faulty component must be in the path between $C$ and $F$ (of course, there
may be other faulty components, but  we are only searching for the
abnormality that explains the observation). We only need to consider the
descriptions of components $y$ and $z$ in order to find the diagnosis.

In the next section we will show how to use the framework described in 
\cite{Frontiers} in order to find diagnoses without having to check
the entire system description for consistency. 

\section{Local Kernel Diagnosis}


As we have seen, diagnosis problems fit very well in the framework for local
change that we proposed in \cite{Local} and \cite{Frontiers}. Besides the fact
that the traditional method for finding diagnosis based on the notion of
consistency is almost identical to the construction of kernel semi-revision,
in most diagnosis problems there is a very natural notion of relatedness
that can be used to structure the belief base so that the search for diagnoses
becomes more efficient.

In this section we formalize the example in Figure \ref{fig7} in order to show
how to derive a concrete relatedness relation from the given database.

\begin{figure}[h]
\vspace*{6cm}
\hspace*{-1cm}
\begin{picture}(100,100)

 \put(30,30){\mbox{\makebox(100,100)[bl]{\includegraphics{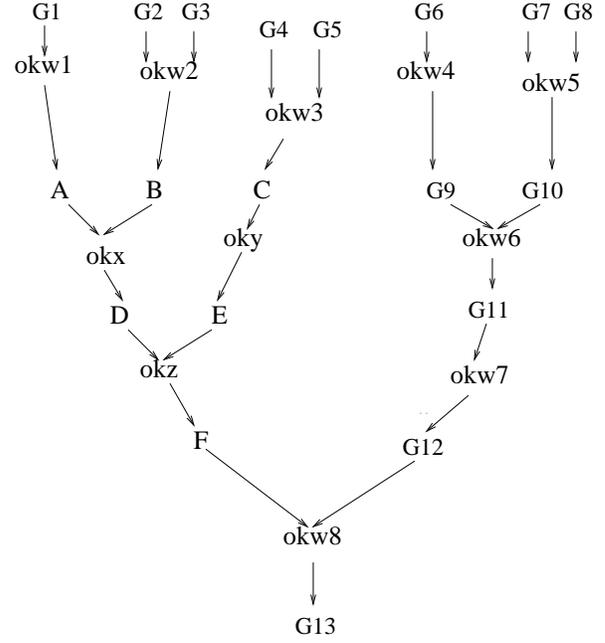}}}}

\end{picture}
\caption{Relatedness relation between atoms}
\label{fig8}
\end{figure}

We will use a relatedness relation between atoms, as illustrated in Figure
\ref{fig8}. The relation is not symmetric. We can easily adapt the definitions presented in
\cite{Frontiers} to deal with a directed graph.

The basic algorithm is as follows: we start from the propositional variables
that occur in the observation and spread the activation in the graph,
following the direction of  the arcs. The spreading finishes either when the
end of the paths are reached or when we run out of resources (time or
memory). This is done by the algorithm {\tt Retrieve} below, an adaptation of
the algorithm given in \cite{Frontiers}. The algorithm uses the function 
{\tt Adjacent} to collect all nodes related to a given node, i.e., given a 
relatedness relation $R$, {\tt Adjacent($x$)}=$\{y\in $Var(SD)$\cup $ASS$|R(x,y)\}$,
where Var($X$) is the set of propositional variables ocurring in the formulas 
of set $X$. For a set $Y$ of propositional variables, {\tt Adjacent}($Y$)=
$\bigcup\{${\tt Adjacent}($y$)$|y\in Y\}$.

\bigskip

{\tt Retrieve}(OBS,ASS,Relevant):

1. For all $p\in$ Var(OBS), mark($p$)

2. $\Delta^1$(OBS) := Adjacent(Var(OBS))

3. Relevant := Var(OBS)$\cap$ASS

4. $i$ := 1;
   stop := false

5. While not stop do

\hspace*{1cm} 5.1. For all $p\in\Delta^i$(OBS), mark($p$)

\hspace*{1.7cm} If $p\in$ASS, 

\hspace*{2.4cm} then Relevant := Relevant $\cup\{p\}$

\hspace*{1cm} 5.2. i := i+1; $\Delta^i$(OBS)=$\emptyset$

\hspace*{1cm} 5.3 For all $p\in\Delta^{i-1}$(OBS),

\hspace*{1.7cm} $\Delta^i$(OBS) := $\Delta^i$(OBS) $\cup
\{q\in$Adjacent($p$) 

\hspace*{2.4cm} s.t. not marked$(q)\}$

\hspace*{1cm} 5.4 If $\Delta^i$(OBS)$=\emptyset$, then stop := true

\bigskip 

After we have retrieved the relevant assumables, the relevant compartment 
is taken to be the observation together with all formulas in SD$\cup$ASS which
mention the relevant assumables.

\bigskip

{\tt Compartment}(OBS,SD,ASS,Comp):

1. Retrieve(OBS,ASS,Relevant)

2. Comp=OBS

3. For all $p\in$Relevant, 

\hspace*{1cm} Comp:= Comp$\cup\{\alpha\in$SD$\cup$ASS$|p\in$Var($\alpha$)$\}$.

\bigskip 

As we have seen in \cite{Frontiers}, the algorithm for {\tt Retrieve} is
an anytime algorithm. The algorithm for {\tt Compartment} is not, at least in
principle. But if one keeps the order in which the relevant atoms are
retrieved and uses them in this order in line 3 of algorithm {\tt Compartment},
one can be sure that the description of the most relevant components will be
retrieved  first.

For the circuit in Figure \ref{fig7}, we have:

\medskip

SD=\{$(A\wedge B)\wedge okX\rightarrow D$,
$\neg(A\wedge B)\wedge okX\rightarrow\neg D$,
  
$C\wedge okY\rightarrow \neg E$,
$\neg C\wedge okY\rightarrow E$,

$(D\vee E)\wedge okZ\rightarrow F$,
$\neg(D\vee E)\wedge okZ\rightarrow \neg F$,

$G1\wedge okW1\rightarrow\neg A$,
$\neg G1\wedge okW1\rightarrow A$,

$(G2\vee G3)\wedge okW2\rightarrow B$,
$\neg (G2\vee G3)\wedge okW2\rightarrow \neg B$,

$(G4\wedge G5)\wedge okW3\rightarrow C$,
$\neg(G4\wedge G5)\wedge okW3\rightarrow\neg C$,

$G6\wedge okW4\rightarrow\neg G9$,
$\neg G6\wedge okW4\rightarrow G9$,

$(G7\wedge G8)\wedge okW5\rightarrow G10$,

$\neg (G7\wedge G8)\wedge okW5\rightarrow\neg G10$,

$(G9\vee G10)\wedge okW6\rightarrow G11$,

$\neg (G9\vee G10)\wedge okW6\rightarrow\neg G11$,

$G11\wedge okW7\rightarrow G12$,
$\neg G11\wedge okW7\rightarrow\neg  G12$,

$(F\wedge G12)\wedge okW8\rightarrow G13$,

$\neg(F\wedge G12)\wedge okW8\rightarrow\neg G13$\}

\smallskip

ASS =\{$okX, okY, okZ, okW1, okW2, okW3, okW4,$

\hspace*{1.5cm}$okW5, okW6, okW7, okW8$\}

\medskip

If we apply the algorithm {\tt Retrieve}($\neg C\wedge\neg F$,ASS,Relevant) to the
graph depicted in Figure \ref{fig8}, we get Relevant=\{okY,okZ,okW8\}.
For {\tt Compartment}(OBS, SD, ASS, Comp) we get 

Comp=$\{\neg C\wedge\neg F, C\wedge okY\rightarrow \neg E,$ 

$\neg C\wedge okY\rightarrow E, (D\vee E)\wedge okZ\rightarrow F,$ 

$\neg(D\vee E)\wedge okZ\rightarrow \neg F,$ 

$(F\wedge G12)\wedge okW8\rightarrow G13,$ 

$\neg(F\wedge G12)\wedge okW8\rightarrow\neg G13, okY, okZ, okW8\}$.

The diagnosis can be searched using only the formulas in Comp.
Note that the component w8 was not really relevant for the diagnosis but,
nevertheless, we have reduced the set to be semi-revised.

This is a very general method for focusing on a small part of the system
description. One can add to it some domain specific heuristics to improve its
efficiency.  The system IDEA \cite{IDEA}, used by FIAT repair centers works on
dependence graphs that show graphically the relation between the several
components of a device.

In \cite{KR00} we have shown that Reiter's algorithm for consistency-based
diagnosis can be used for kernel semi-revision.  
The algorithm for kernel operations can be easily combined with the algorithm
{\tt Compartment} presented in this section.

Applying Reiter's algorithm to Comp, given the observation $\neg C\wedge\neg
F$, we get as possible diagnoses: $\{okY\}$ and $\{okZ\}$.

\section{Conclusions}

In this paper we have shown how to combine Reiter's algorithm for
consistency-based diagnosis with the algorithm for finding the relevant
compartment of a database. The result is a method for finding diagnosis which
focuses on the relevant part of the system description. 

Making clear the similarities between diagnosis and belief revision can be
very profitable for both areas of research. As shown in \cite{KR00}, the
computational tools developed in the field of diagnosis can be adapted to be
used for belief revision. And as we show in this paper, theories developed 
for belief revision can be applied on diagnosis for obtaining more 
efficient methods.

Future work includes the study of other approaches to diagnosis as
well as the study of the computational complexity of the method proposed.

\bibliographystyle{aaaim}
\bibliography{model}

\begin{thebibliography}{}

\bibitem[\protect\citeauthoryear{Alchourr\'on \&
  Makinson}{1985}]{Alchourron_Makinson2}
Alchourr\'on, C., and Makinson, D.
\newblock 1985.
\newblock On the logic of theory change: Safe contraction.
\newblock {\em Studia Logica} 44:405--422.

\bibitem[\protect\citeauthoryear{Hamscher, Console, \& de
  Kleer}{1992}]{Readings_Diagnosis}
Hamscher, W.; Console, L.; and de~Kleer, J., eds.
\newblock 1992.
\newblock {\em Readings in Model-Based Diagnosis}.
\newblock Morgan Kaufmann.

\bibitem[\protect\citeauthoryear{Hansson \& Wassermann}{1999}]{Local}
Hansson, S.~O., and Wassermann, R.
\newblock 1999.
\newblock Local change.
\newblock In preparation (a preliminary version appeared in the Fourth
  Symposium on Logical Formalizations of Commonsense Reasoning, London, 1998).

\bibitem[\protect\citeauthoryear{Hansson}{1994}]{Hansson_Kernel}
Hansson, S.~O.
\newblock 1994.
\newblock Kernel contraction.
\newblock {\em Journal of Symbolic Logic} 59:845--859.

\bibitem[\protect\citeauthoryear{Hansson}{1997}]{Hansson_semi}
Hansson, S.~O.
\newblock 1997.
\newblock Semi-revision.
\newblock {\em Journal of Applied Non-Classical Logic} 7(1-2):151--175.

\bibitem[\protect\citeauthoryear{Reiter}{1987}]{Reiter}
Reiter, R.
\newblock 1987.
\newblock A theory of diagnosis from first principles.
\newblock {\em Artificial Intelligence} 32:57--95.
\newblock Reprinted in \cite{Readings_Diagnosis}.

\bibitem[\protect\citeauthoryear{Sanseverino \& Cascio}{1997}]{IDEA}
Sanseverino, M., and Cascio, F.
\newblock 1997.
\newblock Model-based diagnosis for automotive repair.
\newblock {\em IEEE Expert} 12(6):33--37.

\bibitem[\protect\citeauthoryear{Wassermann}{1999}]{Frontiers}
Wassermann, R.
\newblock 1999.
\newblock On structured belief bases.
\newblock In Rott, H., and Williams, M.-A., eds., {\em Frontiers in Belief
  Revision}. Kluwer.
\newblock to appear.

\bibitem[\protect\citeauthoryear{Wassermann}{2000}]{KR00}
Wassermann, R.
\newblock 2000.
\newblock An algorithm for belief revision.
\newblock In {\em Proceedings of the Seventh International Conference on
  Principles of Knowledge Representation and Reasoning (KR2000)}.
\newblock Morgan Kaufmann.

\end{thebibliography}

\end{document}